\documentclass{article}

\usepackage{microtype}
\usepackage{graphicx}
\usepackage{subfigure}
\usepackage{booktabs} 

\usepackage{hyperref}

\usepackage{tikz} 
\usepackage{pgfplots}
\pgfplotsset{compat=1.8}
 \usetikzlibrary{
  angles,
 	arrows,
 	arrows.meta,
 	shapes,
 	shapes.symbols,
 	shapes.callouts,
 	shapes.geometric,
 	arrows,
 	matrix,
 	backgrounds,
 	positioning,
 	plotmarks,
 	calc,
 	patterns,
 	matrix,
 	decorations.pathreplacing,
 	decorations.pathmorphing,
 	decorations.text,
 	decorations.shapes,
 	decorations.fractals,
 	decorations.markings,
 	spy,
  quotes
 }
\usepgfplotslibrary{fillbetween}
\usepgfplotslibrary{statistics}
\usepgfplotslibrary{groupplots}
\usepgfplotslibrary{units}
\pgfplotsset{
only if/.style args={entry of #1 is #2}{
/pgfplots/boxplot/data filter/.code={
\edef\tempa{\thisrow{#1}}
\edef\tempb{#2}
\ifx\tempa\tempb
\else
\def\pgfmathresult{}
\fi
}
}
}

\pgfplotsset{
  /pgfplots/confidence box/.style 2 args={
    legend image code/.code={
        \definecolor{steelblue31119180}{RGB}{31,119,180}
        \draw[steelblue31119180,no markers, fill=steelblue31119180, opacity=0.5]
        plot coordinates {
        (-0.1cm,-0.1cm)
        (-0.1cm,0.2cm)
        (0.5cm,0.2cm)
        (0.5cm,-0.1cm)
        (-0.1cm,-0.1cm)
      }
      node[rectangle]{};
    }
  }
}

\usepackage[acronym]{glossaries}
\glsdisablehyper

\usepackage{collcell}
\usepackage{hhline}
\usepackage{multirow}

\usepackage{lscape}

\usetikzlibrary{external}
\def\colorModel{hsb} 

\newcommand\ColCell[1]{
  \pgfmathparse{#1<50?1:0}  
    \ifnum\pgfmathresult=0\relax\color{white}\fi
    \pgfmathsetmacro\compA{0.6666-#1/150} 
    \pgfmathsetmacro\compB{1}             
    \pgfmathsetmacro\compC{1}             
  \edef\x{\noexpand\centering\noexpand\cellcolor[\colorModel]{\compA,\compB,\compC}}\x #1
  } 
\newcolumntype{E}{>{\collectcell\ColCell}m{0.4cm}<{\endcollectcell}}  

 \usepackage[accepted]{icml2024}

\usepackage{amsmath}
\usepackage{amssymb}
\usepackage{mathtools}
\usepackage{amsthm}

\usepackage[capitalize,noabbrev]{cleveref}

\theoremstyle{plain}
\newtheorem{theorem}{Theorem}[section]

\newtheorem{lemma}[theorem]{Lemma}
\newtheorem{corollary}[theorem]{Corollary}
\theoremstyle{definition}

\theoremstyle{remark}

\usepackage[disable,textsize=tiny]{todonotes}

\newacronym[plural=AIC]{aic}{AIC}{Akaike Information Criterium}
\newacronym{aks}{AKS}{Adjusting Kernel Search}

\newacronym[plural=AIC]{bic}{BIC}{Bayesian Information Criterium}

\newacronym{cks}{CKS}{Compositional Kernel Search}

\newacronym[plural=GPs,firstplural=Gaussian Processes (GPs)]{gp}{GP}{Gaussian process}

\newacronym{hmc}{HMC}{Hamiltonian Monte Carlo}

\newacronym{kl}{KL}{Kullback-Leibler}
\newacronym[plural=LODE-GPs,firstplural=Linear Ordinary Differential Equation \glspl{gp} (LODE-GPs)]{lodegp}{LODE-GP}{Linear Ordinary Differential Equation \gls{gp}}

\newacronym{map}{MAP}{Maximum A Posteriori}
\newacronym{mc}{MC}{Monte Carlo}
\newacronym{mcmc}{MCMC}{Markov Chain Monte Carlo}
\newacronym{ml}{ML}{Machine Learning}
\newacronym{mll}{MLL}{Marginal Log Likelihood}

\newacronym{nuts}{NUTS}{No U-Turn Sampling}

\newacronym{rmse}{RMSE}{Root Mean Squared Error}

\newacronym{skc}{SKC}{Scalable Kernel Composition}

\usepackage{multirow}

\newcommand{\MLL}{\ensuremath{p(y|X,\theta)}}

\DeclareMathOperator*{\Lap}{\text{Lap}}

\DeclareMathOperator*{\LapS}{\text{Lap}_0}
\DeclareMathOperator*{\LapAIC}{\text{Lap}_A}
\DeclareMathOperator*{\LapBIC}{\text{Lap}_B}

\begin{document}

\twocolumn[
\icmltitle{On the Laplace Approximation as Model Selection Criterion for Gaussian Processes}



\icmlsetsymbol{equal}{*}

\begin{icmlauthorlist}
\icmlauthor{Andreas Besginow}{th}
\icmlauthor{Jan David Hüwel}{fhag}
\icmlauthor{Thomas Pawellek}{}
\icmlauthor{Christian Beecks}{fhag}
\icmlauthor{Markus Lange-Hegermann}{th}
\end{icmlauthorlist}

\icmlaffiliation{th}{University of Applied Sciences and Arts, Lemgo, Germany}
\icmlaffiliation{fhag}{University of Hagen, Hagen, Germany}

\icmlcorrespondingauthor{Andreas Besginow}{andreas.besginow@th-owl.de}

\icmlkeywords{Machine Learning, ICML}

\vskip 0.3in
]



\printAffiliationsAndNotice{}  

\begin{abstract}
Model selection aims to find the best model in terms of accuracy, interpretability or simplicity, preferably all at once.
In this work, we focus on evaluating model performance of Gaussian process models, i.e.\ finding a metric that provides the best trade-off between all those criteria.
While previous work considers metrics like the likelihood, AIC or dynamic nested sampling, they either lack performance or have significant runtime issues, which severely limits applicability.
We address these challenges by introducing multiple metrics based on the Laplace approximation, where we overcome a severe inconsistency occuring during naive application of the Laplace approximation.
Experiments show that our metrics are comparable in quality to the gold standard dynamic nested sampling without compromising for computational speed.
Our model selection criteria allow significantly faster and high quality model selection of Gaussian process models.
\end{abstract}

\newcommand{\textLap}{Laplace}
\newcommand{\textLapS}{stabilized Laplace}
\newcommand{\textLapAIC}{\gls{aic} corrected Laplace}
\newcommand{\textLapBIC}{\gls{bic} corrected Laplace}

	\section{Introduction}

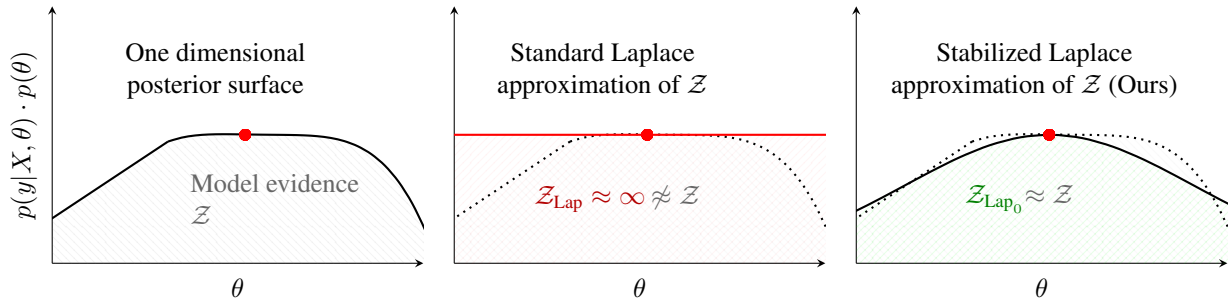
\begin{figure*}\label{fig:introduction_conceptual_laplace_pathology}
    \begin{center}
    \begin{tikzpicture}
        \begin{axis}[
            axis lines = left,
            xlabel = $\theta$,
            ylabel = {$p(y | X, \theta)\cdot p(\theta)$},
            ymin = -0,
            ymax = 2,
            xmin = -1.5,
            xmax = 1.0,
            ytick=\empty,
            xtick=\empty,
            height=5cm,
            width=0.38\textwidth,
            legend pos=north east,
        ]
            \addplot [domain=-0.72:1.473, smooth, thick, name path=A] {-0.5*x^4-0.2*x^3 -0.01*x+1};
            \addplot [thick, name path=B] plot [smooth] coordinates {(-0.72, 0.9474) (-3.5, -1.182)};

            \addplot [draw=none, name path=nil, domain=-3.5:1.473] {-2};

            \addplot [pattern=north west lines, pattern color= gray, opacity=0.3] fill between[of=A and nil,soft clip={domain=-0.73:1.473}]; 
            \addplot [pattern=north west lines, pattern color= gray, opacity=0.3] fill between[of=B and nil,soft clip={domain=-3.5:-0.72}]; 

            \addplot [mark=*, color=red] (-0.2, 1);
            \node[align=left] at (axis cs:0, 0.5) {\textcolor{gray!80!black}{Model evidence} \\ \textcolor{gray!80!black}{$\mathcal{Z}$}};
            \node[align=center] at (axis cs:-0.4, 1.5) {One dimensional \\ posterior surface};
        \end{axis}
    \end{tikzpicture}
    \begin{tikzpicture}
        \begin{axis}[
            axis lines = left,
            xlabel = $\theta$,
            ytick=\empty,
            xtick=\empty,
            ymin = -0,
            ymax = 2,
            xmin = -1.5,
            xmax = 1.0,
            height=5cm,
            width=0.38\textwidth,
            legend pos=north east,
        ]
            \addplot [dotted,domain=-0.72:1.473, smooth, thick, name path=A] {-0.5*x^4-0.2*x^3 -0.01*x+1};
            \addplot [dotted,thick, name path=B] plot [smooth] coordinates {(-0.72, 0.9474) (-3.5, -1.182)};
            \addplot [draw=none, name path=nil, domain=-3.5:1.473] {-2};
            \addplot [pattern=north west lines, pattern color= gray, opacity=0.2] fill between[of=A and nil,soft clip={domain=-0.73:1.473}]; 
            \addplot [pattern=north west lines, pattern color= gray, opacity=0.2] fill between[of=B and nil,soft clip={domain=-3.5:-0.72}]; 

            \addplot [color=red, thick, name path=lap_app, domain=-3.5:1.473] {1};
            \addplot [pattern=north east lines, pattern color= red, opacity=0.2] fill between[of=lap_app and nil,soft clip={domain=-3.5:1.473}]; 
            \addplot [mark=*, color=red] (-0.2, 1);
            \node at (axis cs:-0.4, 0.5) {\color{red!70!black} $\mathcal{Z}_{\text{Lap}} \approx \infty \textcolor{gray!80!black}{\:\not\approx \mathcal{Z}}$};
            \node[align=center] at (axis cs:-0.5, 1.5) {Standard Laplace \\approximation of $\mathcal{Z}$};
        \end{axis}
    \end{tikzpicture}
    \begin{tikzpicture}
        \begin{axis}[
            axis lines = left,
            xlabel = $\theta$,
            ymin = -0,
            ymax = 2,
            xmin = -1.5,
            xmax = 1.0,
            ytick=\empty,
            xtick=\empty,
            height=5cm,
            width=0.38\textwidth,
            legend pos=north east,
        ]

            \addplot [dotted,domain=-0.72:1.473, smooth, thick, name path=A] {-0.5*x^4-0.2*x^3 -0.01*x+1};
            \addplot [dotted,thick, name path=B] plot [smooth] coordinates {(-0.72, 0.9474) (-3.5, -1.182)};

            \addplot [draw=none, name path=nil, domain=-3.5:1.473] {-2};

            \addplot [pattern=north west lines, pattern color= gray, opacity=0.2] fill between[of=A and nil,soft clip={domain=-0.73:1.473}]; 
            \addplot [pattern=north west lines, pattern color= gray, opacity=0.2] fill between[of=B and nil,soft clip={domain=-3.5:-0.72}]; 

            \addplot [domain=-3.5:1.473, smooth, thick, name path=lap_app] {exp(-0.5*((x+0.2)^2*1.0592))};
            \addplot [pattern=north east lines, pattern color= green, opacity=0.3] fill between[of=lap_app and nil,soft clip={domain=-3.5:1.473}]; 
            \addplot [mark=*, color=red] (-0.2, 1);
            \node at (axis cs:-0.4, 0.5) {\color{green!50!black}$\mathcal{Z}_{\LapS} \textcolor{gray!80!black}{\approx \mathcal{Z}}$};
            \node[align=center] at (axis cs:-0.3, 1.5) {Stabilized Laplace \\ approximation of $\mathcal{Z}$ (Ours)};
        \end{axis}
    \end{tikzpicture}
    \end{center}
    \caption{
    A conceptual visualization of the inconsistency when naively applying the Laplace approximation and one of our suggested variants.
    \textbf{Left}: The posterior over parametrizations $\theta$, with a degenerate local extremum (red dot). The model evidence $\mathcal{Z}$ is the gray shaded area.
    \textbf{Middle}: Naive application of the Laplace approximation around the optimum with infinitely large model evidence approximation $\mathcal{Z}_{\text{Lap}} \approx \infty$ overlaid in red.
    \textbf{Right}: Application of our \textLapS{} ($\LapS$) around the optimum with model evidence approximation $\mathcal{Z}_{\LapS} \approx \mathcal{Z}$ overlaid in green.
    }
\end{figure*}
Turning data into knowledge frequently requires inferring descriptive models which capture the major data characteristics.
In particular in the domain of data science (and engineering), a common challenge lies in identifying an appropriate descriptive \gls{ml} model that effectively learns interpretable information based on a given dataset without overfitting and underfitting.
 
When faced with small datasets, \glspl{gp} have become the preferred method for flexibly and interpretably modeling complex patterns in regression tasks.
The interpretability and transparency offered by \glspl{gp} can be attributed to the selection of a covariance function (also often called kernels), which possesses the ability to capture a wide array of structures arising from diverse domains such as geometry \cite{borovitskiy2020matern}, symmetry \cite{holderrieth2021equivariant}, harmonic analysis \cite{lazaro20sparse}, or differential equations \cite{besginow2022constraining,alvarez2009latent,harkonen2022gaussian}.
These covariance functions induce a strong prior that result in reasonable models even in the case of scarce data.

In cases where no such prior knowledge can be induced for covariance functions, one can resort to kernel search algorithms \cite{duvenaud2013structure,berns2022automated,huwel2021automated} whose goal is selecting the best covariance function w.r.t.\ a given performance measure.
These methods generally follow an iterative process to construct a descriptive covariance function from a set of base kernels and operations for a given regression data set. 
To ensure a good fit between model and data, kernel searches utilize different performance measures to evaluate the constructed models \cite{duvenaud2013structure,kim2018scaling}.
In the interest of interpretability, covariance functions with a minimal number of operands (e.g.\ summands) are generally preferred.

Most covariance functions come with (hyper)parameters.
The values of hyperparameters are usually determined reliably by maximizing the \gls{mll} $\log p(y|X,\theta)$ of the Gaussian likelihood \cite{rasmussen2006gaussian}.
So, inside of a single parametrized class of kernels, it is possible to identify the most suitable \emph{parametrization} for a model.
However, identifying the most suitable \emph{class of kernels} purely from data remains a challenging task.

The gold standard for \gls{gp} model selection is the computation of the model evidence $\mathcal{Z} = p(y|X)=\int p(y|X,\theta)p(\theta)\operatorname{d}\!\theta$, which necessitates the challenging task of marginalizing over the hyperparameters $\theta$.
Achieving a high-quality approximation of this integral requires computationally intensive methods (cf.\ Figure~\ref{fig:kernelsearch_runtimes}) such as dynamic nested sampling \cite{skilling2006nested}.
However, to enhance computational efficiency, more approximate techniques are often employed, such as the \gls{mll}, the \gls{aic} or the \gls{bic}.
Though, while the \gls{mll} is commonly utilized for hyperparameter fitting, it tends to overfit when used as a model selection criterion, as it never penalizes superfluous parameters. 
To address this limitation, \gls{aic} penalizes the number of parameters added to the \gls{mll}, thus mitigating overfitting.
Alternatively, \gls{bic} aims to recover the original model and therefore imposes a larger penalty dependent on both the number of parameters and the number of data points \cite{burnham2004multimodel,burnham1998practical}.
Both of these approximations only take the number of hyperparameters and data points into account, and hence are rather crude.

In this study, we introduce a novel collection of model selection criteria for \glspl{gp} that are not only computationally efficient but also yield robust performance.
These criteria derive from the Laplace approximation of the parameter posterior to compute the model evidence integral \cite{bishop2006pattern,jaynes2003probability}.
While the Laplace approximation has been used to train \gls{gp} hyperparameters or estimate the posterior predictive distribution \cite{li2023improving,flaxman2015fast,kuss2005assessing,zilber2021vecchia,hartmann2019laplace}, to the best of our knowledge, this paper is the first to use Laplace approximation to compare the \gls{gp} model evidence \emph{between model classes}.

For this use case, we identify severe inconsistencies when naively applying the Laplace approximation where adding superfluous parameters improves the model evidence.
To address these inconsistencies, we introduce variants of the Laplace approximation, by bounding the eigenvalues of the Hessian.
Using our variants of the Laplace approximations prevents that additional parameters contribute positively to the model evidence, without significant contributions to the model fit.
We illustrate the extreme case of these inconsistencies in Figure~\ref{fig:introduction_conceptual_laplace_pathology}, where we mitigate the infinitely large approximation of the model evidence $\mathcal{Z}$, due to naive application of the standard Laplace approximation, using our Laplace approximations.

We show in experiments that our Laplace approximations perform as good as dynamic nested sampling while retaining a small runtime.
Additionally, in kernel search experiments we show how our different Laplace approximations have different strengths with respect to test performance and recognition of the underlying model.

The core contributions of this paper are as follows\footnote{Code will be published on acceptance}.
\begin{itemize}
    \item We introduce new model selection criteria, based on the Laplace approximation, which mitigate the original inconsistency coming with naive application of the Laplace approximation
    \item We show that our criteria are comparable in approximation of the model evidence to dynamic nested sampling and offer similar interpretability as dynamic nested sampling, all while having a neglectable runtime
    \item In kernel search experiments we show the superiority of our Laplace approximations in predicting the model evidence of models, compared to the state of the art
\end{itemize}

	\section{Preliminaries}

\subsection{Gaussian Processes}
A \glsfirst{gp} $g = \mathcal{GP}(\mu, k)$ is a stochastic process where every finite set of realizations at the points $x_i$ are jointly Gaussian \cite{rasmussen2006gaussian}.
Such a \gls{gp} is characterized by its mean function $\mu : \mathbb{R}^d \to \mathbb{R} : x \mapsto \mathbb{E}(g(x))$ (often set to zero) and its positive semi-definite covariance function (or kernel) $k : \mathbb{R}^d \times \mathbb{R}^d \to \mathbb{R} : (x, x') \mapsto \mathbb{E}((g(x) - \mu(x))(g(x') - \mu(x'))^T)$.
Most covariance functions contain various hyperparameters, e.g.\ the Squared Exponential (SE) kernel $k_{SE}(x, x') = \sigma^2_f \exp(\frac{-(x - x')^2}{2\ell^2})$ contains signal variance $\sigma_f$ and lengthscale $\ell$, which become part of the \gls{gp} $g_{SE} = \mathcal{GP}(0, k_{SE})$.

To train the \gls{gp} hyperparameters, the \glsfirst{mll} is the de facto \gls{gp} loss function, defined as:
\begin{equation}\label{eq:mll}
    \begin{aligned}
        \log \MLL = &- \frac{1}{2}{y}^T(K + \sigma_n^2I)^{-1}{y} - \frac{1}{2}\log|K + \sigma_n^2I| \\
                    &- \frac{n}{2}\log(2\pi)  
    \end{aligned}
\end{equation}
where $X \in \mathbb{R}^{n\times d}$ and $y \in \mathbb{R}^n$ are a dataset of size $n$ and the \glspl{gp} hyperparameters $\theta$ are hidden in the calculation of the covariance matrix $K$ of observations $X$.

We refer to all possible parametrizations of \glspl{gp} with a specific kernel as a \emph{class} of \glspl{gp}, e.g.\ all parametrizations of $g_{SE}$ form the class of \glspl{gp} with the SE kernel.

The set of kernels is closed under various operations \cite{duvenaud2014automatic,rasmussen2006gaussian,jidling2017linearly}. 
Among such operations are addition and multiplication of kernels, for example combining a periodic kernel with an squared exponential kernel as $k = k_{PER} \cdot k_{SE}$ enforces locally periodic and smooth behaviour.
This grows the set of potentially useful classes of \glspl{gp} significantly, making model selection a complex endevaour.
\subsection{Model Selection for GPs}\label{sec:related_work_model_selection}
Model selection tries to find the best model for a dataset ($y, X$) in terms of accuracy, interpretability or simplicity, preferably all at once.
A general rule when selecting models is preferring simpler models that explain the data just good enough, often referred to as Occam's razor.
This should be reflected in model selection criteria, which weigh model performance versus complexity.

Naively, one can use the optimized \gls{mll} $\widehat{\mathcal{L}} = \log p(y|X,\hat{\theta}) $ to assess the model quality for \gls{gp} $g$ with optimal hyperparameters $\hat{\theta}$.
This comes with the downside that increased model complexity allows to (over)fit the data easier and any additional hyperparameter will never decrease the \gls{mll} value.
Additionally we are solely dependent on the optimization procedure, which might get caught in ``bad'' optima (cf.\ Section~\ref{sec:evaluation_mauna_loa_dataset}).
Finally, the \gls{mll} is not appropriate for model evaluation under specific conditions, e.g.\ for $m$-constant mean functions as discussed in \cite{karvonen2023maximum}.

The performance of a class of \glspl{gp} can be measured objectively for a dataset through the model evidence (sometimes called marginalized likelihood), i.e.\ the likelihood marginalized over the hyperparameters $\theta$:
\begin{equation}\label{eq:gp_likelihood_integral}
    \mathcal{Z} = p(y|X) = \int p(y|X,\theta)p(\theta)d\theta
\end{equation} 
Since this integral is intractable one usually resorts to approximations.
The most precise approximation to this integral is through sampling based approaches like nested sampling \cite{skilling2006nested}.
At its core, the likelihood surface $\MLL$ is explored by repeatedly sampling hyperparameters from a prior $p(\theta)$.
By calculating the weighted sum of those likelihood evaluations, we get a good approximation of the likelihood integral i.e.\ model evidence $\mathcal{Z}$.
But performing this calculation is often computationally infeasible, even for \glspl{gp} with two hyperparameters and ten datapoints we need several minutes for this approximation.

Due to this, we look into faster approximations of the likelihood.
The earliest is \gls{aic}~\cite{akaike1974new}, which is based on the optimized \gls{mll} $\widehat{\mathcal{L}}$ and corrects the optimum for the number of hyperparameters $u$ as follows: $\text{AIC} = 2u - 2\widehat{\mathcal{L}}$.
It has its roots in information theory and ``is derived from a frequentist framework, and cannot be interpreted as an approximation to the model evidence'' (cf.\ p.\ 162 in \cite{murphy2012machine}). 
Related to \gls{aic} is \gls{bic}, which changes the correction term in dependence to the number of datapoints as $\text{BIC} = u\log(n) - 2\widehat{\mathcal{L}}$ \cite{schwarz1978estimating}, and can be considered a simple approximation to the log model evidence (cf.\ §5.2.5.1 in \cite{murphy2022probabilistic}).
However, as stated in p.\ 33 of \cite{bishop2006pattern}: ``Such criteria do not take account of the uncertainty in the model hyperparameters, however, and in practice they tend to favour overly simple models.''

Most of the methods were applied in various works \cite{simpson2021marginalised,kristiadi2021learnable,green2015bayesian,ritter2018online,duvenaud2013structure,lloyd2014automatic,huwel2021automated,harkonen2022gaussian} to approximate different measures with respect to \glspl{gp}.
For example, \cite{simpson2021marginalised} apply nested sampling to sample from the hyperparameter posterior or \cite{duvenaud2013structure} improve kernel search through \gls{bic}.

\subsection{Kernel Search Algorithms}
The quality and interpretability of a \gls{gp} model depends strongly on the chosen kernel \cite{lloyd2014automatic,rasmussen2006gaussian}.
While there are commonly used kernels, such as the squared exponential kernel, which works well for generally smooth data, other kernels can provide a better fit for specific kinds of data and allow the inclusion of prior knowledge into the modelling process \cite{duvenaud2014automatic}.
For example, the periodic and the linear kernel are optimal, if data exhibits periodic and linear behavior, respectively. 

In situations without prior knowledge about the data, an automated kernel search algorithm can infer a fitting kernel from the data in an iterative selection process.
Introduced with \gls{cks} in \cite{duvenaud2013structure,lloyd2014automatic}, such algorithms construct kernels in an iterative, greedy fashion based on a collection of ``base kernels'' and operations (e.g.\ $+$, $\cdot$) w.r.t.\ some given performance measure, usually the \gls{mll} \cite{huwel2021automated} or \gls{bic} \cite{duvenaud2013structure,lloyd2014automatic}.
Many adaptations of those algorithms exist nowadays \cite{kim2018scaling,berns20213cs,berns2022automated,huwel2022dynamically}. 
Alternative approaches replace the iterative process by learning spectral distributions and model them as a mixture of Gaussians \cite{li2019implicit,wilson2013gaussian}.

In our experiments, we compare the commonly used performance measures \gls{mll} and \gls{bic} with \gls{aic}, \gls{map} and our Laplace approximations to show their applicability in kernel search.

	\section{Laplace approximation of GP model evidence}\label{sec:laplaceApprox}
In this section, we detail how to apply the Laplace approximation to approximately solve the model evidence integral in Equation~\eqref{eq:gp_likelihood_integral}.
The Laplace approximation $\Lap(f)$ aims to find a Gaussian approximation for a function $f$ using the second order Taylor approximation in $\log$-space of $f$ around its optimum, i.e.\ a point of gradient equal to zero.

The Laplace approximation has been used in numerous works to train \gls{gp} hyperparameters or use it to integrate out the latent \gls{gp} $f$, instead of the parameters $\theta$, \cite{li2023improving,flaxman2015fast,kuss2005assessing,zilber2021vecchia,hartmann2019laplace}.
Notable is its applicability to non-Gaussian likelihoods e.g.\ to optimize hyperparameters for classification \glspl{gp} and approximate the corresponding marginal likelihood (see chapters 3.4 and 3.5 in \cite{rasmussen2006gaussian}).

In this work we use the Laplace approximation with the goal of computing the model evidence integral in formula~\eqref{eq:gp_likelihood_integral}.
Approximating the product $p(y | X, \theta)\cdot p(\theta)$ of likelihood and prior by a second order Taylor approximation in log-space gives us the following:
\begin{equation}\label{eq:gp_likelihood_prior_laplace_approximation}
    \begin{aligned}
        \log \Lap(p(y | X, \theta)\cdot p(\theta)) \approx &\log (p(y | X, \hat{\theta})\cdot p(\hat{\theta})) \\
                                                     &- \frac{1}{2}{(\theta - \hat{\theta})}^TH(\theta - \hat{\theta})
    \end{aligned}
\end{equation}
Here, $H = -\nabla \nabla \log(p(y | X, \theta)\cdot p(\theta))|_{\theta = \hat{\theta}}$ ist the Hessian at a (local) optimum $\hat{\theta}$ \cite{bishop2006pattern}, which is easily computed with the autodiff functionality in most \gls{ml} libraries.

We use the Formula~\eqref{eq:gp_likelihood_prior_laplace_approximation} to approximate the integral, resulting in the log model evidence $\log\mathcal{Z}$:
\begin{align}\label{eq:model_evidence_integral_approximated_by_laplace_approximation}
    \log\mathcal{Z} \approx \log\mathcal{Z}_{\text{Lap}} = &\log\left(p(y | X, \hat{\theta})\cdot p(\hat{\theta})\right) + \frac{u}{2}\log(2\pi) \\
    &- \frac{1}{2}\log(|H|)
\end{align}
where $\hat{\theta}$ is the optimum found during optimization, $u$ is the number of hyperparameters and $H$ is the Hessian at $\hat{\theta}$.
For details of this derivation we refer to Appendix~\ref{app:solution_model_evidence_integral_with_laplace_approximation}.

Sadly, the next subsection shows that this standard form of Laplace approximation, is not suitable to approximate the model evidence of \glspl{gp}.

\subsection{Overcoming inconsistencies of the Laplace approximation}\label{subsection_pathologies}
Naively applying the standard Laplace approximation $\log\mathcal{Z}_{\text{Lap}}$ in formula~\eqref{eq:model_evidence_integral_approximated_by_laplace_approximation} exhibits an inconsistency, preventing its practical use.
In the following we discuss how this inconsistency causes the standard Laplace approximation to increase the model evidence by adding superfluous hyperparameters without necessarily contributing to the model fit.
In extreme cases this inconsistency increases the model evidence $\mathcal{Z}_{\text{Lap}}$ to infinity, as illustrated in Figure~\ref{fig:introduction_conceptual_laplace_pathology} for one dimension.
There, a local extremum that is degenerate, in the sense that it is very weak, causes the standard Laplace approximation to result in an infinite model evidence $\mathcal{Z}_{\text{Lap}}$.
This directly contradicts Occam's razor and prevents the usage of the Laplace approximation for approximating the model evidence.

To address this problem, we introduce various versions of the Laplace approximation, all with different interpretations.
Doing so, we ensure that the mere presence of a hyperparameter does not contribute to an arbitrary improvement.

The inconsistencies are directly connected to the Hessian $H$ and are best interpreted in one dimension ($u=1$) after disentangling the dimensions by diagonalizing the Hessian.
For hyperparameters where the, one dimensional, Hessian $H=\begin{bmatrix}\lambda\end{bmatrix}$ is moderately small (e.g.\ for $\lambda < 1$), we can directly observe from formula~\eqref{eq:model_evidence_integral_approximated_by_laplace_approximation} that $\frac{1}{2}\log(2\pi) - \frac{1}{2}\log(\lambda)$ has a \emph{positive} contribution to the model evidence just by the hyperparameter of $\lambda$ \emph{existing}, even without a contribution to the model fit.
We therefore modify the Hessian matrix to mitigate this behaviour, similar to Newton methods with Hessian modification in second order optimization (cf.\ Section 3.4 in \cite{nocedal1999numerical}).

\begin{lemma}\label{lem:laplace_approx_min_neg_contribution}
    To ensure that each hyperparameter of a \gls{gp} has a minimal negative contribution $r$ in the last to summands of formula~\eqref{eq:gp_likelihood_prior_laplace_approximation} to the model evidence $\mathcal{Z}$, every eigenvalue $\lambda$ of the Hessian needs to be at least:
    \begin{equation}\label{eq:hessian_prior_correction}
        \lambda \geq \exp(-2r)\cdot 2\pi
    \end{equation}
\end{lemma}
For the proof we refer the reader to Appendix~\ref{app:laplace_minimal_negative_correction_derivation}.

Based on this equation we suggest three improved variants of the Laplace approximation: \emph{\textLapS} ($\LapS$), \emph{\textLapAIC} ($\LapAIC$) and \emph{\textLapBIC} ($\LapBIC$).
\begin{corollary}\label{cor:Lap0_corollary}
    Stabilized Laplace ($\LapS$):
    To ensure a minimal negative contribution $r=0$ to the log model evidence $\log\mathcal{Z}_{\LapS} \approx \log\mathcal{Z}$ every eigenvalue has to be at least $\lambda \geq 2\pi$.
\end{corollary}
\begin{corollary}\label{cor:LapA_corollary}
    \textLapAIC{} ($\LapAIC$):
    To ensure a minimal negative contribution $r=-1$ to the model evidence $\log\mathcal{Z}_{\LapAIC} \approx \log\mathcal{Z}$ every eigenvalue has to be at least $\lambda \geq \exp(2)\cdot 2\pi$.
\end{corollary}
\begin{corollary}\label{cor:LapB_corollary}
    \textLapBIC{} ($\LapBIC$):
    To ensure a minimal negative contribution $r=-\log(n)$ to the model evidence $\log\mathcal{Z}_{\LapBIC} \approx \log\mathcal{Z}$ every eigenvalue has to be at least $\lambda \geq \exp(2\log(n))\cdot 2\pi=2\pi\cdot n^2$.
\end{corollary}

Each of the variants has a different interpretation. 
Stabilized Laplace ($\LapS$) prevents additional hyperparameters from improving the model evidence without a significant contribution to the model fit.
\textLapAIC{} ($\LapAIC$) corrects the optimization result similar to \gls{aic} by contributing $r=-1$ for each hyperparameter.
Finally, \textLapBIC{} ($\LapBIC$) is correcting similar to \gls{bic} by contributing $r=-\log(n)$ for each hyperparameter.
The connection to \gls{aic} resp.\ \gls{bic} derives from the fact that these corrected Laplace approximations collapse into  \gls{aic} resp.\ \gls{bic} in the case of all eigenvalues $\lambda$ of the Hessian being small (see appendix \ref{app:Lap_collapse_proof}).

These lower bounds for the eigenvalues $\lambda$ of the Hessian $H$ prevents the inconsistencies model evidence for \gls{gp} likelihoods.
Non-Gaussian likelihoods, also log concave ones like the logistic likelihood in p.\ 42 in \cite{rasmussen2006gaussian}, could also benefit from a similar eigenvalue correction since e.g.\ log concavity doesn't necessarily prevent infinitely large (or small) likelihood values from superfluous hyperparameters.

	\section{Evaluation}
\definecolor{darkgray176}{RGB}{176,176,176}
\definecolor{gray}{RGB}{128,128,128}

\definecolor{claret}{RGB}{108, 6, 37}
\definecolor{cocoabrown}{RGB}{210, 106, 37}
\definecolor{pink}{RGB}{255,192,203}
\definecolor{purple}{RGB}{128,0,128}
\definecolor{turquoise}{RGB}{64,224,208}
We demonstrate that our improved variants of the Laplace approximation are superior model selection criteria by comparing them to both the state of the art model selection metrics \gls{aic} and \gls{bic} and also to \gls{mll} and \gls{map}.
We use dynamic nested sampling \cite{higson2019dynamic} to approximate the model evidence, with a precision of up to two decimal places, as our ground truth and perform the following experiments:
\begin{enumerate}
    \item A comparison of all metrics on a small dataset for a single \gls{gp} to explore the detailed behaviour of the Laplace approximation and point out the benefits for all metrics in an interpretable way.
    \item A kernel search experiment where we vary the selection criteria.
	Here, the large scale evaluation of models ensures the reproducibility of our results.
    \item Comparing all metrics on the Mauna Loa dataset for increasingly more complex kernels to show its applicability on a large and complex real world dataset
\end{enumerate}

Additional details regarding the experiments can be found in their respective appendices under Appendix~\ref{app:additional_experiment_details}.
We also detail the normal distributed prior for the hyperparameters we use for nested sampling and our variants of the Laplace approximation in Appendix~\ref{app:prior_definition}.

\subsection{Interpretable example}\label{sec:evaluation_interpretable_example}
\begin{figure}
    \input{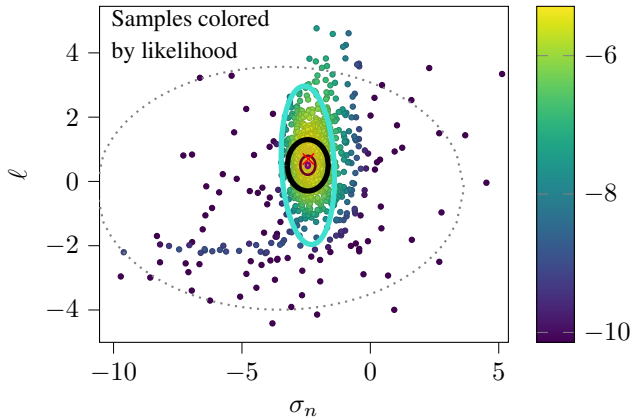}
    \definecolor{darkgray176}{RGB}{176,176,176}
    \definecolor{gray}{RGB}{128,128,128}
    \definecolor{pink}{RGB}{255,192,203}
    \definecolor{cocoabrown}{RGB}{210, 106, 37}
    \definecolor{purple}{RGB}{128,0,128}
    \definecolor{turquoise}{RGB}{64,224,208}
    \caption{
    Results of a nested sampling for the linear noisy dataset in Section~\ref{sec:evaluation_interpretable_example} (showing 1024 representative samples out of 12,255 total samples), higher values are better.
    The differently colored ellipses show the $2\sigma$ confidence ellipses for the normal distributions associated with the Laplace approximations for $\LapS$ (black), $\LapAIC$ (\textcolor{claret}{brown}), $\LapBIC$ (\textcolor{purple}{purple}) and the standard Laplace approximation (\textcolor{turquoise}{blue}).
    The dotted \textcolor{darkgray176}{gray} ellipse is the $2\sigma$ confidence ellipse for the hyperparameter prior.
    The \textcolor{red}{$\times$} shows the optimum found during nested sampling.
    We see that the ellipses derived from our versions of the Laplace approximation give a good approximation of the most relevant area of the likelihood surface, since they cover the majority of the samples of nested sampling.
    }
    \label{fig:linear_noise_SE_nested_sampling}
\end{figure}

We demonstrate the performance of our Laplace approximations and their similarity to the gold standard, nested sampling, through a detailed exploration of the model evidence and the relevant area of the likelihood surface.
We do this for an evenly spaced dataset drawn from a linear function with small noise, and apply an SE-\gls{gp} containing hyperparameters lengthscale $\ell$ and noise $\sigma_n$ (for details see Appendix~\ref{app:additional_experiment_interpretable_experiment_details}).

Figure~\ref{fig:linear_noise_SE_nested_sampling} shows the (our versions and standard) Laplace approximations and the samples used to approximate the log model evidence $\log\mathcal{Z}$ during nested sampling.
These samples can be interpreted as approximately coming from the hyperparameter posterior.
And since we approximate the hyperparameter posterior using the Laplace approximation we can conclude that a high overlap \emph{of the samples and the Laplace approximations} indicates a high overlap \emph{of the relevant area of the likelihood surface}.
And using the Gaussian interpretation of the Laplace approximation to draw their $2\sigma$ confidence ellipses, we can see exactly that.
The colored ellipses for the different variants of the Laplace approximation contain 78.4\% (standard Laplace in \textcolor{turquoise}{blue}), 60.6\% ($\LapS$ in black), 22.8\% ($\LapAIC$ in \textcolor{claret}{brown}) and 3.1\% ($\LapBIC$ in \textcolor{purple}{purple}) of the samples (cf.\ Figure~\ref{fig:linear_noise_SE_nested_sampling}).
As \textLapS{} ($\LapS$) encircles the majority of the highest valued points, this again underlines that our variants of the Laplace approximation take the most relevant area of the likelihood surface into account for our approximation.
By using stricter negative contributions in our Laplace approximations the corresponding $2\sigma$ confidence ellipses for the normal distributions naturally contain fewer and fewer points, which is clearly visible when comparing \textLapAIC{} ($\LapAIC$, \textcolor{claret}{brown}) or \textLapBIC{} ($\LapBIC$, \textcolor{purple}{purple}) to \textLapS{} ($\LapS$, \textcolor{black}{black}).

In addition to the visual results of Figure~\ref{fig:linear_noise_SE_nested_sampling} showing the similarities of nested sampling and our Laplace approximations, they also show numerical similarities.
Nested sampling approximates the log model evidence as $\log\mathcal{Z} = -8.12$ and \textLapS{} ($\LapS$) as $\log\mathcal{Z}_{\LapS} = -9.17$.
The stricter negative contributions in our Laplace approximations reflect in their log model evidence approximations where \textLapAIC{} ($\LapAIC$) approximates $\log\mathcal{Z}_{\LapAIC} = -11.17$ and \textLapBIC{} ($\LapBIC$) as $\log\mathcal{Z}_{\LapBIC} = -13.78$.
Interestingly \gls{aic} and \gls{bic} also provide good results, when rescaling them by $-0.5$ to live on the same scale as the model evidence, where \gls{aic} approximates $\log\mathcal{Z}_{AIC} = -7.285$ and \gls{bic} $\log\mathcal{Z}_{BIC} = -7.59$. 
Even though \gls{aic} and \gls{bic} are similarly close to $\log\mathcal{Z}$, they lack any interpretation comparable to our confidence ellipses and don't allow a diagnosis of overfitting hyperparameters or whether they actually represent the majority of the accepted samples.

\subsection{Kernel search experiments}\label{sec:evaluation_kernel_search}
\begin{table}[t]
    \caption{The ratio of times the data generating kernel was recognized when performing \gls{cks} with the respective metric, varied over number of datapoints $N$. We see that on average, our variants $\LapS$, $\LapAIC$ and $\LapBIC$ are strong model selection criteria.}
    \label{tab:kernelsearch_recognition_ratio}
    \vskip 0.15in
    \begin{center}
    \begin{small}
    \begin{sc}
                
    \begin{tabular}{lrrrrrrr}
        \toprule
        $N$ &	MLL & AIC &	BIC & MAP &	$\LapS$ & $\LapAIC$ & $\LapBIC$ \\
        \midrule
        5   & 2.5 & 42.5 & \textbf{45.0} & 0.0 & 42.5 & 42.5 & \textbf{45.0} \\ 
        10  & 2.5 & 47.5 & \textbf{52.5} & 0.0 & 47.5 & 50.0 & 47.5 \\ 
        20  & 0.0 & \textbf{65.0} & 60.0 & 0.0 & 55.0 & 55.0 & 55.0 \\ 
        30  & 0.0 & 60.0 & 60.0 & 0.0 & 55.0 & 62.5 & \textbf{65.0} \\ 
        40  & 0.0 & 60.0 & 60.0 & 0.0 & 62.5 & \textbf{67.5} & 65.0 \\ 
        50  & 0.0 & 57.5 & 65.0 & 0.0 & 62.5 & \textbf{72.5} & 67.5 \\ 
        100 & 0.0 & 60.0 & 70.0 & 0.0 & 67.5 & \textbf{75.0} & 72.5 \\ 
        200 & 0.0 & 70.0 & 65.0 & 0.0 & \textbf{75.0} & 70.0 & 70.0 \\ 
        avg.\ & 0.6 & 57.8 & 59.7 & 0.0 & 58.4 & \textbf{61.9} & 60.9\\ 

        \bottomrule
    \end{tabular}
    \end{sc}
    \end{small}
    \end{center}
    \vskip -0.1in
\end{table}

\begin{table*}[t]
    \caption{The average log model evidence $\log\mathcal{Z}$, approximated using nested sampling, $\pm$ one standard deviation for models selected during the kernel search experiment.
    The columns show the results when using the respective metric as the performance measure in \gls{cks}.
    The rows show the number of training datapoints used.
    Higher values are better and the best performing run w.r.t.\ the model evidence is bold.
    The results show that using \textLapS{} ($\LapS$) as the target metric results in kernels with better model evidence.
    }
    \label{tab:kernelsearch_model_evidence}
    \vskip 0.15in
    \begin{center}
    \begin{small}
    \begin{sc}
    \begin{tabular}{lrrrrrrr}
        \toprule
        $N$ & MLL & AIC & BIC & MAP & $\LapS$ & $\LapAIC$ & $\LapBIC$\\
        \midrule
        5 & $-9.00 \pm 2.56$ & $-5.38 \pm 3.71$ & $\mathbf{-5.28 \pm 3.85}$ & $-8.80 \pm 2.69$ & $-5.83 \pm 4.45$ & $-5.84 \pm 4.43$ & $-5.87 \pm 4.38$\\
        10 & $-14.12 \pm 6.42$ & $-5.37 \pm 5.82$ & $-5.72 \pm 5.91$ & $-13.90 \pm 6.40$ & $\mathbf{-5.32 \pm 6.56}$ & $-5.46 \pm 6.56$ & $-5.63 \pm 6.71$\\
        20 & $-23.50 \pm 15.63$ & $-2.54 \pm 8.94$ & $-3.54 \pm 8.52$ & $-23.92 \pm 15.90$ & $\mathbf{-1.84 \pm 8.91}$ & $-2.03 \pm 9.04$ & $-2.51 \pm 9.12$\\
        30 & $-33.30 \pm 26.26$ & $\mathbf{5.25 \pm 11.29}$ & $4.53 \pm 11.45$ & $-33.32 \pm 25.28$ & $4.11 \pm 11.03$ & $3.84 \pm 11.22$ & $3.39 \pm 11.37$\\
        40 & $-45.48 \pm 37.89$ & $12.33 \pm 12.01$ & $12.09 \pm 12.03$ & $-40.77 \pm 36.68$ & $\mathbf{13.32 \pm 11.56}$ & $13.09 \pm 11.63$ & $12.85 \pm 11.78$\\
        50 & $-49.42 \pm 48.85$ & $21.14 \pm 14.19$ & $\mathbf{21.29 \pm 14.54}$ & $-48.06 \pm 47.07$ & $18.35 \pm 12.82$ & $18.02 \pm 12.97$ & $17.79 \pm 13.10$\\
        100 & $-94.41 \pm 100.15$ & $57.92 \pm 16.73$ & $57.95 \pm 17.10$ & $-89.25 \pm 96.13$ & $\mathbf{59.57 \pm 18.63}$ & $59.50 \pm 18.73$ & $59.14 \pm 18.73$\\
        200 & $-163.60 \pm 196.11$ & $145.34 \pm 19.33$ & $146.40 \pm 20.86$ & $-189.39 \pm 207.50$ & $\mathbf{147.13 \pm 20.76}$ & $147.06 \pm 20.78$ & $146.66 \pm 20.63$\\ 
        \bottomrule
    \end{tabular}
    \end{sc}
    \end{small}
    \end{center}
\end{table*}
\begin{table*}[t]
    \caption{The average log likelihood (normalized by number of datapoints $N$) on a test dataset $\pm$ one standard deviation for models selected during the kernel search experiment, averaged over the ten test datasets.
    The test datasets are drawn from the same distributions used to generate the kernel selection datasets.
    The columns show the results when using the respective metric as the performance measure in \gls{cks}.
    The rows show the number of training datapoints used.
    Higher values are better and the best performing run w.r.t.\ the test dataset is bold.
    The best performing runs are mostly shared between \gls{bic} and our \textLapBIC{} ($\LapBIC$).
    }
    \label{tab:kernelsearch_test_likelihood}
    \vskip 0.15in
    \begin{center}
        \begin{small}
            \begin{sc}

    \begin{tabular}{lrrrrrrr}
        \toprule
        $N$ & MLL & AIC & BIC & MAP & $\LapS$ & $\LapAIC$ & $\LapBIC$\\
        \midrule
        5 & $-1.53 \pm 0.56$ & $-3.09 \pm 3.67$ & $-2.92 \pm 3.48$ & $\mathbf{-1.47 \pm 0.51}$ & $-1.76 \pm 1.57$ & $-1.76 \pm 1.57$ & $-1.76 \pm 1.57$\\
        10 & $-1.35 \pm 0.65$ & $-0.82 \pm 1.06$ & $\mathbf{-0.81 \pm 1.00}$ & $-1.22 \pm 0.78$ & $-1.79 \pm 3.44$ & $-1.31 \pm 1.98$ & $-1.15 \pm 1.70$\\
        20 & $-1.34 \pm 0.92$ & $-0.10 \pm 0.72$ & $\mathbf{0.02 \pm 0.55}$ & $-1.29 \pm 1.05$ & $-3.01 \pm 12.41$ & $-2.82 \pm 12.39$ & $-0.59 \pm 3.51$\\
        30 & $-1.10 \pm 0.88$ & $0.20 \pm 0.33$ & $0.21 \pm 0.33$ & $-1.05 \pm 0.88$ & $-0.30 \pm 2.45$ & $-0.31 \pm 2.45$ & $\mathbf{0.21 \pm 0.40}$\\
        40 & $-1.16 \pm 1.04$ & $\mathbf{0.34 \pm 0.28}$ & $\mathbf{0.34 \pm 0.28}$ & $-1.46 \pm 1.92$ & $0.23 \pm 0.75$ & $0.33 \pm 0.31$ & $0.32 \pm 0.31$\\
        50 & $-1.46 \pm 2.68$ & $-0.12 \pm 2.61$ & $-0.12 \pm 2.64$ & $-1.21 \pm 2.21$ & $-0.16 \pm 3.35$ & $-0.16 \pm 3.35$ & $\mathbf{-0.09 \pm 3.34}$\\
        100 & $-1.14 \pm 1.16$ & $0.03 \pm 3.70$ & $-0.26 \pm 4.14$ & $-1.16 \pm 1.24$ & $-0.31 \pm 4.05$ & $-0.31 \pm 4.05$ & $\mathbf{0.58 \pm 0.30}$\\
        200 & $-0.97 \pm 1.04$ & $0.44 \pm 1.48$ & $0.22 \pm 1.89$ & $-0.95 \pm 1.03$ & $0.22 \pm 1.87$ & $0.22 \pm 1.87$ & $\mathbf{0.73 \pm 0.09}$\\
        \bottomrule

    \end{tabular}
    \end{sc}
    \end{small}

    \end{center}
    \vskip -0.1in
\end{table*}

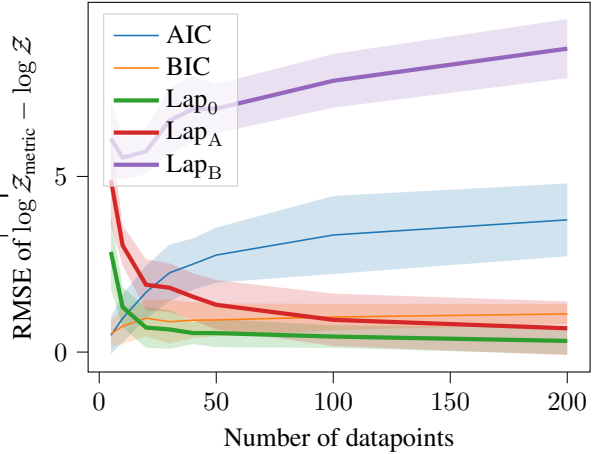
\begin{figure}
\begin{tikzpicture}

\definecolor{crimson2143940}{RGB}{214,39,40}
\definecolor{darkgray176}{RGB}{176,176,176}
\definecolor{darkorange25512714}{RGB}{255,127,14}
\definecolor{forestgreen4416044}{RGB}{44,160,44}
\definecolor{lightgray204}{RGB}{204,204,204}
\definecolor{mediumpurple148103189}{RGB}{148,103,189}
\definecolor{steelblue31119180}{RGB}{31,119,180}

\begin{axis}[
legend cell align={left},
legend style={
  fill opacity=0.7,
  draw opacity=1,
  text opacity=1,
  at={(0.03,0.97)},
  anchor=north west,
  draw=lightgray204
},
tick align=outside,
tick pos=left,
x grid style={darkgray176},
xmin=-4.75, xmax=209.75,
height=6.5cm,
width=\columnwidth,
xtick style={color=black},
xlabel={Number of datapoints},
ylabel={RMSE of $\log\mathcal{Z}_{\text{metric}} - \log\mathcal{Z}$},
y grid style={darkgray176},
ymin=-0.567366501688957, ymax=9.95602675378323,
ytick style={color=black}
]
\path [fill=steelblue31119180, fill opacity=0.2]
(axis cs:5,1.00994038581848)
--(axis cs:5,-0.0548948347568512)
--(axis cs:10,0.272113561630249)
--(axis cs:20,0.950867056846619)
--(axis cs:30,1.450066447258)
--(axis cs:40,1.76977121829987)
--(axis cs:50,1.97751808166504)
--(axis cs:100,2.22096800804138)
--(axis cs:200,2.72584295272827)
--(axis cs:200,4.80129671096802)
--(axis cs:200,4.80129671096802)
--(axis cs:100,4.44426822662354)
--(axis cs:50,3.54519414901733)
--(axis cs:40,3.23606061935425)
--(axis cs:30,3.06073141098022)
--(axis cs:20,2.453697681427)
--(axis cs:10,1.64617669582367)
--(axis cs:5,1.00994038581848)
--cycle;

\path [fill=darkorange25512714, fill opacity=0.2]
(axis cs:5,0.881932973861694)
--(axis cs:5,0.156223744153976)
--(axis cs:10,0.224022090435028)
--(axis cs:20,0.4423907995224)
--(axis cs:30,0.245783269405365)
--(axis cs:40,0.394387125968933)
--(axis cs:50,0.445717513561249)
--(axis cs:100,0.621431231498718)
--(axis cs:200,0.803672671318054)
--(axis cs:200,1.36884200572968)
--(axis cs:200,1.36884200572968)
--(axis cs:100,1.3739185333252)
--(axis cs:50,1.3852207660675)
--(axis cs:40,1.41829645633698)
--(axis cs:30,1.48423361778259)
--(axis cs:20,1.48785972595215)
--(axis cs:10,1.24875664710999)
--(axis cs:5,0.881932973861694)
--cycle;

\path [fill=forestgreen4416044, fill opacity=0.2]
(axis cs:5,3.89622688293457)
--(axis cs:5,1.80752766132355)
--(axis cs:10,0.70243513584137)
--(axis cs:20,0.122171938419342)
--(axis cs:30,0.0925111174583435)
--(axis cs:40,0.227221697568893)
--(axis cs:50,0.137000679969788)
--(axis cs:100,0.119064211845398)
--(axis cs:200,-0.0658417046070099)
--(axis cs:200,0.702709794044495)
--(axis cs:200,0.702709794044495)
--(axis cs:100,0.763334453105927)
--(axis cs:50,0.942957878112793)
--(axis cs:40,0.856660723686218)
--(axis cs:30,1.19779634475708)
--(axis cs:20,1.27163219451904)
--(axis cs:10,1.85348665714264)
--(axis cs:5,3.89622688293457)
--cycle;

\path [fill=crimson2143940, fill opacity=0.2]
(axis cs:5,6.17356014251709)
--(axis cs:5,3.62525749206543)
--(axis cs:10,2.46025276184082)
--(axis cs:20,1.18665814399719)
--(axis cs:30,1.14025521278381)
--(axis cs:40,0.906531929969788)
--(axis cs:50,0.642014563083649)
--(axis cs:100,0.172486066818237)
--(axis cs:200,-0.0890304446220398)
--(axis cs:200,1.44283676147461)
--(axis cs:200,1.44283676147461)
--(axis cs:100,1.66375970840454)
--(axis cs:50,2.04966378211975)
--(axis cs:40,2.24669313430786)
--(axis cs:30,2.52504825592041)
--(axis cs:20,2.64474296569824)
--(axis cs:10,3.61506271362305)
--(axis cs:5,6.17356014251709)
--cycle;

\path [fill=mediumpurple148103189, fill opacity=0.2]
(axis cs:5,7.19326972961426)
--(axis cs:5,4.9586706161499)
--(axis cs:10,4.9207124710083)
--(axis cs:20,5.0635929107666)
--(axis cs:30,5.64779138565063)
--(axis cs:40,5.97307443618774)
--(axis cs:50,6.22654962539673)
--(axis cs:100,6.95849704742432)
--(axis cs:200,7.79756307601929)
--(axis cs:200,9.47769069671631)
--(axis cs:200,9.47769069671631)
--(axis cs:100,8.48727321624756)
--(axis cs:50,7.62878656387329)
--(axis cs:40,7.8595929145813)
--(axis cs:30,7.52966260910034)
--(axis cs:20,6.35949802398682)
--(axis cs:10,6.14448928833008)
--(axis cs:5,7.19326972961426)
--cycle;

\addplot [semithick, steelblue31119180]
table {%
5 0.477522750411204
10 0.959145143888173
20 1.70228225553305
30 2.25539900324701
40 2.50291582655481
50 2.7613560430883
100 3.33261805910153
200 3.76356980261489
};
\addlegendentry{AIC}
\addplot [semithick, darkorange25512714]
table {%
5 0.519078346824641
10 0.736389421244444
20 0.965125251697107
30 0.86500843633668
40 0.906341819080159
50 0.915469177901909
100 0.99767486696734
200 1.08625739001981
};
\addlegendentry{BIC}
\addplot [ultra thick, forestgreen4416044]
table {%
5 2.8518772464179
10 1.27796092062534
20 0.696902027920418
30 0.645153780825442
40 0.541941197358465
50 0.53997925428666
100 0.441199327151084
200 0.318434033363624
};
\addlegendentry{$\LapS$}
\addplot [ultra thick, crimson2143940]
table {%
5 4.89940898296964
10 3.03765781141005
20 1.91570054132198
30 1.83265175646167
40 1.57661249222901
50 1.34583915716343
100 0.918122858706473
200 0.676903170042191
};
\addlegendentry{$\LapAIC$}
\addplot [ultra thick, mediumpurple148103189]
table {%
5 6.07597011063947
10 5.53260095046733
20 5.71154556913759
30 6.58872690715535
40 6.91633367939746
50 6.92766787828509
100 7.72288498298291
200 8.63762634115299
};
\addlegendentry{$\LapBIC$}
\end{axis}

\end{tikzpicture}
    \caption{The \gls{rmse} $\pm$ one standard deviation, between the log model evidence and the respective metric's value, across varying dataset sizes.
    \gls{aic} and \gls{bic} have been rescaled by $-0.5$ to have the same scale as the model evidence.
    Smaller \gls{rmse} is better.
    Our variants of the Laplace approximation are drawn in bold.}
    \label{fig:kernel_search_model_evidence_to_metric_value_difference}
\end{figure}

In this experiment we perform the \glsfirst{cks}, varying its performance measure between \gls{mll}, \gls{map}, \gls{aic}, \gls{bic} or our three variants of the Laplace approximations.
We generate datasets of varying sizes by sampling ten times from four different \glspl{gp}, for a total of 40 different datasets per dataset size (more details in Appendix~\ref{app:additional_experiment_kernel_search_experiment_details}).
For each dataset-metric combination we perform \gls{cks} for, at most, three iterations i.e.\ resulting kernels contain at most three base kernels.
Nested sampling could not be used as a performance measure for \gls{cks} due to the large computation time required per model evaluation.

Our variant \textLapS{} ($\LapS$) outperforms the state of the art in its approximation to the model evidence.
This is shown in Figure~\ref{fig:kernel_search_model_evidence_to_metric_value_difference} where our approximation to the log model evidence $\log\mathcal{Z}_{\LapS}$ reaches an increasingly small distance to the log model evidence $\log\mathcal{Z}$.
This distance increases for \gls{aic} and \gls{bic} for bigger data set sizes, making them unsuitable in practice.

Since \textLapS{} ($\LapS$) approximates $\log\mathcal{Z}$ well, the resulting kernels from the kernel search also have the highest log model evidence, as clearly shown in Table~\ref{tab:kernelsearch_model_evidence}.
We outperform the state of the art in the important task of finding the kernels with the highest model evidence.

The log likelihoods in Table~\ref{tab:kernelsearch_test_likelihood} indicate that our \textLapBIC{} ($\LapBIC$) exceeds at finding models that work for \emph{different datasets from the same distribution}.
This table shows the log likelihood $\log p(y_*|X_*,\theta)$ on a \emph{test dataset} $(X_*,y_*)$, which is \emph{not} the predictive \gls{gp} likelihood $p(y_* | X_*, y, X, \theta)$, but was calculated by replacing the training datapoints $(X,y)$ with the test datapoints $(y_*, X_*)$ sampled from the same prior \gls{gp} as the training data.
Hence, these resulting kernels represent the original distribution and are flexible enough to model new data from this distribution.

Finally, we show that \textLapAIC{} ($\LapAIC$) is best at recognizing the underlying data generating model as shown in Table~\ref{tab:kernelsearch_recognition_ratio}.
The state of the art, though comparable, is mostly outperformed and we clearly see how \gls{mll} and \gls{map} tend to result in models of maximal size and thus fail most clearly to recognize the data generating kernels.

\begin{figure}
\begin{tikzpicture}

\definecolor{crimson2143940}{RGB}{214,39,40}
\definecolor{darkgray176}{RGB}{176,176,176}
\definecolor{darkorange25512714}{RGB}{255,127,14}
\definecolor{forestgreen4416044}{RGB}{44,160,44}
\definecolor{gray127}{RGB}{127,127,127}
\definecolor{lightgray204}{RGB}{204,204,204}
\definecolor{mediumpurple148103189}{RGB}{148,103,189}
\definecolor{orchid227119194}{RGB}{227,119,194}
\definecolor{sienna1408675}{RGB}{140,86,75}
\definecolor{steelblue31119180}{RGB}{31,119,180}

\begin{axis}[
legend cell align={left},
legend style={
  fill opacity=0.8,
  draw opacity=1,
  text opacity=1,
  at={(0.98,0.48)},
  anchor=east,
  draw=lightgray204
},
log basis y={10},
tick align=outside,
tick pos=left,
x grid style={darkgray176},
xlabel={Number of datapoints},
xmin=-4.75, xmax=209.75,
xtick style={color=black},
y grid style={darkgray176},
ylabel={$\log$ runtime in seconds},
ymin=0.777713193907485, ymax=346.328892238016,
ymode=log,
ytick style={color=black},
height=5cm,
width=\columnwidth,
ytick={0.01,0.1,1,10,100,1000,10000},
yticklabels={
  \(\displaystyle {10^{-2}}\),
  \(\displaystyle {10^{-1}}\),
  \(\displaystyle {10^{0}}\),
  \(\displaystyle {10^{1}}\),
  \(\displaystyle {10^{2}}\),
  \(\displaystyle {10^{3}}\),
  \(\displaystyle {10^{4}}\)
}
]
\addplot [semithick, gray127]
table {%
5 61.1851812342803
10 77.7637485027313
20 98.9885983943939
30 105.606123505036
40 104.134088607629
50 92.4854519049327
100 76.3236302912235
200 88.0557717521985
};
\addlegendentry{Nested}
\addplot [ultra thick, forestgreen4416044]
table {%
5 1.13184404029991
10 1.33589151122353
20 1.29205522194053
30 1.29362973643072
40 1.503276069959
50 1.77419417030884
100 1.39923956466444
200 1.68840442534649
};
\addlegendentry{$\LapS$}
\addplot [ultra thick, crimson2143940]
table {%
5 1.10046248056672
10 1.3481835038373
20 1.48628099958102
30 1.43653705752257
40 1.31073676492229
50 1.62464626449527
100 1.3311915818489
200 1.74246676799023
};
\addlegendentry{$\LapAIC$}
\addplot [ultra thick, mediumpurple148103189]
table {%
5 1.02615631927143
10 1.34313613555648
20 1.26286904450619
30 1.28384616031791
40 1.28434641794725
50 1.59900390397419
100 1.33740224296396
200 1.80936512621966
};
\addlegendentry{$\LapBIC$}
\end{axis}

\end{tikzpicture}
    \caption{Average time to calculate our metrics, averaged over each kernel evaluated during kernel searches, on a logarithmic scale.
    For our metrics, the variants of the Laplace approximation, the time includes the training procedure with two random restarts.
    This shows that the computation time of our approaches are two orders of magnitude smaller than that of dynamic nested sampling.
    }
    \label{fig:kernelsearch_runtimes}
\end{figure}
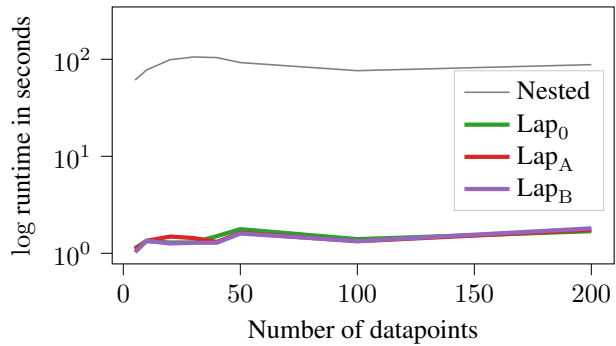

These results show how for varying goals, different negative contributions and thus different variants of our Laplace approximations are best fit: $\LapS$ for approximating the model evidence, $\LapAIC$ for the current dataset, and $\LapBIC$ for extrapolation quality.
All this while being two orders of magnitude faster than dynamic nested sampling (cf.\ Figure~\ref{fig:kernelsearch_runtimes}).

Finally, we emphasize the importance of our introduced variants.
Here, we just focus on the extreme case of the inconsistency leading to infinite model evidences as in Figure~\ref{fig:introduction_conceptual_laplace_pathology}.
We observed $\log\mathcal{Z}_{\text{Lap}} \approx \pm \infty$, $40.35$\% of the time.
Out of $4884$ kernels that were tested during the kernel search, $1971$ approximations to the model evidence were completely meaningless.
Hence, only our correction to the Laplace approximation made this kernel search experiment possible.

\subsection{Real world dataset}\label{sec:evaluation_mauna_loa_dataset}

\begin{table}[t]
    \caption{Mauna Loa performance for \gls{mll}, \gls{map}, \gls{aic}, \gls{bic}, our Laplace approximations and the approximated log model evidence $\log\mathcal{Z}$, according to dynamic nested sampling, for the increasingly more sophisticated kernels $k_n$.
    We rescale \gls{aic} and \gls{bic} with $-0.5$ to be on the same scale as the log model evidence and therefore call them $\log\mathcal{Z}_{AIC}$ and $\log\mathcal{Z}_{BIC}$, here.
    The approximated log model evidences $\log\mathcal{Z}$ for kernels $k_{1}$ through $k_{1+2+3+4}$ have error estimates of $\pm 3.78$, $\pm 3.441$, $\pm 0.29$ and $\pm 5.19$.
    Higher values indicate that the models are evaluated as better performing.
    }
    \label{tab:mauna_loa_metrics}
    \vskip 0.15in
    \begin{center}
    \begin{small}
    \begin{sc}
    \begin{tabular}{lrrrr}
        \toprule
        $k$       &  $k_{1}$             & $k_{1+2}$  &  $k_{1+2+3}$        &  $k_{1+2+3+4}$ \\
        \midrule
        $\log\mathcal{Z}$    &  $-1258.88$          & $-581.45$  &  $-293.40$          &  $-286.81$\\
        $\LapS$   &  $-1305.48$          & $-655.84$  &  $-288.03$          &  $-288.76$\\
        $\LapAIC$ &  $-1308.10$          & $-660.84$  &  $-295.95$          &  $-298.69$ \\
        $\LapBIC$ &  $-1324.00$          & $-695.87$  &  $-348.29$          &  $-361.64$\\
        $\log\mathcal{Z}_{AIC}$       &  $-1206.85$           & $-554.67$  &  $-143.32$           &  $-144.09$ \\
        $\log\mathcal{Z}_{BIC}$       &  $-1213.30$           & $-569.72$  &  $-164.83$           &  $-169.89$ \\
        MLL       &  $-1203.85$          & $-547.66$  &  $-133.32$          &  $-132.08$\\
        MAP       &  $-1305.10$          & $-651.77$  &  $-285.29$          &  $-286.03$\\                 
        \bottomrule
    \end{tabular}
    \end{sc}
    \end{small}
    \end{center}
    \vskip -0.1in
\end{table}
We conclude the experiments with a discussion of the metrics performance on the Mauna Loa dataset \cite{keeling1994atmospheric}.
We reconstruct the kernel in formula~(5.19) of \cite{rasmussen2006gaussian} by iteratively adding its summands as our tested kernels.
For each such tested kernel we use \gls{aic}, \gls{bic}, \gls{mll}, \gls{map}, our variants of the Laplace approximations and nested sampling to estimate kernel performance.
The summands are $k_1$ (formula~\eqref{eq:rasmussen_launa_k1}) through $k_4$ (formula~\eqref{eq:rasmussen_launa_k4}) and we name our tested kernels $k_{1}$ through $k_{1+2+3+4}$ to indicate which subkernels they consist of.

\begin{align}
    k_1 &= \theta_1^2 \exp\left(-\frac{(x-x')^2}{2\theta_2^2}\right)\label{eq:rasmussen_launa_k1}\displaybreak[0]\\
    k_2 &= \theta_3^2\exp\left(-\frac{(x-x')^2}{2\theta_4^2} - \frac{2\sin^2(\pi (x-x'))}{\theta_5^2}\right)\displaybreak[0]\\
    k_3 &= \theta_6^2\left(1 + \frac{(x - x')^2}{2\theta_8\theta_7^2}\right)^{-\theta_8}\displaybreak[0]\\
    k_4 &= \theta_9^2\exp\left(-\frac{(x-x')^2}{2\theta_{10}^2}\right) + \theta_{11}^2 \delta\label{eq:rasmussen_launa_k4}
\end{align}
where $\delta$ is the Dirac delta function that is $1$ if $x=x'$ and $0$ otherwise.

The results in Table~\ref{tab:mauna_loa_metrics} show that adding more summands improves performance, which can be expected since each summand was crafted to represent a specific aspect of the dataset.
All metrics reflect this behaviour for up to three summands ($k_{1}$ through $k_{1+2+3}$) and also tend to stay very close to the estimated model evidence of nested sampling.
Addition of the fourth component, however, is less valued by all metrics (including the approximate model evidence).

Interestingly, in contrast to previous results in Section~\ref{sec:evaluation_interpretable_example} and for kernels $k_{1}$ and $k_{1+2}$, \gls{aic} and \gls{bic} start to deviate significantly from the approximate log model evidence.
This is apparent from the significant overestimation of the log model evidence in columns $k_{1+2+3}$ and $k_{1+2+3+4}$, compared to $k_{1}$ and $k_{1+2}$.
We can directly explain this through the value of \gls{mll} and the fact that \gls{aic} and \gls{bic} are derived from it.
We see in Table~\ref{tab:mauna_loa_metrics} that \gls{mll} deviates significantly from the log model evidence which directly affects \gls{aic} and \gls{bic}.
We assume this is due to an unstable optimum, i.e.\ declining rapidly when deviating from it, which \gls{mll} optimization found and \gls{aic}/\gls{bic} use to represent the model evidence.
This problem is less pronounced for the \gls{map} optima since they smooth out such unstable optima via the prior.
This stresses the importance of using priors, as is done in our variants of the Laplace approximation.

We conclude that highly complex likelihood surfaces may be at risk of such deviations and suggest using our Laplace approximations there.
In any case, an investigation into detecting such unstable optima would be highly beneficial to make more informed metric choices.
For example, this behaviour wasn't apparent in Section~\ref{sec:evaluation_kernel_search}.
In conclusion this experiment is a clear indicator for the stability of our variants of the Laplace approximation as they do not fail for $k_{1+2+3}$ and $k_{1+2+3+4}$, but further reduce the approximation gap to the log model evidence $\log\mathcal{Z}$.

	\section{Conclusion}

In this paper we introduce novel variants of the Laplace approximation as model selection metrics for \glspl{gp}.
We also provide a method to deal with the inherent inconsistencies of the naive application of the standard Laplace approximation by replacing eigenvalues of the Hessian with varying interpretable thresholds.
We show in experiments that our new variants have comparable performance to dynamic nested sampling, which is considered the gold standard in model evidence approximation and that \textLapS{} ($\LapS$) is a good predictor for the model evidence.
It further outperforms all other metrics in the kernel search experiment w.r.t.\ models with the highest model evidence.
We further show in  that \textLapAIC{} ($\LapAIC$) resp.\ \textLapBIC{} ($\LapBIC$) are best in recognizing the underlying model, used to generate the data, resp.\ finding models that generalize well to new data from the same distribution. 
And finally we demonstrate applicability on real world datasets using the Mauna Loa dataset and highlight a weakness of \gls{aic} and \gls{bic} with respect to optima chosen by \gls{mll}.

\section*{Acknowledgements}
This research was supported by the research training group ``Dataninja'' (Trustworthy AI for Seamless Problem Solving: Next Generation Intelligence Joins Robust Data Analysis) funded by the German federal state of North Rhine-Westphalia.
%
%

\newpage
\section*{Societal Impact}
This paper presents work whose goal is to advance the field of Machine Learning.
There are many potential societal consequences of our work, none which we feel must be specifically highlighted here.

\bibliography{example_paper}
\bibliographystyle{icml2024}

\newpage
\appendix
\onecolumn
\section*{Appendices}
\renewcommand{\thesubsection}{\Alph{subsection}}

\subsection{Additional experiment details}\label{app:additional_experiment_details}
If not otherwise specified we use the following settings in our experiments:

We use GPyTorch's \cite{gardner2018gpytorch} implementation for \glspl{gp}.

\gls{gp} training was performed using PyGRANSO \cite{liang2022ncvx}, a Python port of the GRANSO \cite{curtis2017bfgssqp} optimizer with either \gls{mll} loss for \gls{mll}, \gls{aic} and \gls{bic} or \gls{map} loss for \gls{map} and our Laplace approximations.
The training using PyGRANSO was repeated five times, randomly reinitializing the \gls{gp} every time, for at most 1000 iterations every training.

We estimate our ground truth model evidence $\mathcal{Z}$ with dynamical nested sampling \cite{higson2019dynamic} using the dynesty library \cite{koposiv2023dynesty,speagle2020dynesty} with standard settings, i.e.\ automatic sampler selection, \verb|multi| bounds and automatic live point selection until the the improvement to the model evidence is less than $0.01$ (i.e.\ \verb|dlogz| $<0.01$).
The hyperparameter prior used in nested sampling is constructed from the priors discussed in Appendix~\ref{app:prior_definition}.

\subsubsection{Interpretable experiment}\label{app:additional_experiment_interpretable_experiment_details}
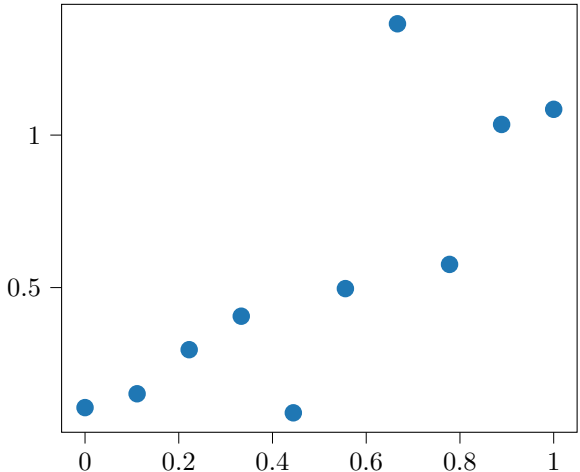
\begin{figure}
    \begin{center}
\begin{tikzpicture}

    \definecolor{darkgray176}{RGB}{176,176,176}
    \definecolor{steelblue31119180}{RGB}{31,119,180}
    
    \begin{axis}[
    tick align=outside,
    tick pos=left,
    x grid style={darkgray176},
    xmin=-0.05, xmax=1.05,
    xtick style={color=black},
    y grid style={darkgray176},
    ymin=0.0255863228175356, ymax=1.42874099998873,
    ytick style={color=black}
    ]
    \addplot [semithick, steelblue31119180, mark=*, mark size=3, mark options={solid}, only marks]
    table {%
    0 0.106470838606225
    0.111111111111111 0.151844221539413
    0.222222222222222 0.296365731603068
    0.333333333333333 0.406171032190665
    0.444444444444444 0.0893660808707719
    0.555555555555556 0.496633373724801
    0.666666666666667 1.3649612419355
    0.777777777777778 0.576025393790963
    0.888888888888889 1.03487772075105
    1 1.08454376768281
    };
    \end{axis}
    
    \end{tikzpicture}
    
    \end{center}
    \caption{The dataset used in the first experiment.
    Ten datapoints $y = x + \sigma_n$ at evenly distributed locations $x = 0\ldots 1$ with $\sigma_n \sim \mathcal{N}(0, 0.1)$}

    \label{fig:appendix_linear_dataset_interpretable_experiment}
\end{figure}

We used the linear dataset shown in Figure~\ref{fig:appendix_linear_dataset_interpretable_experiment}.
The tested \gls{gp} was a standard Squared Eponential (SE) \gls{gp} with an additional noise hyperparameter.

\subsubsection{Kernel search experiment}\label{app:additional_experiment_kernel_search_experiment_details}
In the kernel search experiment we perform \gls{cks} with depth three, i.e.\ kernels can have at most three elements.
The set of base kernels for \gls{cks} contains the SE kernel, the linear kernel and the Matérn $3/2$ kernel.
The kernels are not scaled with additional coefficients $\sigma_f$.
But all \gls{gp} models have a noise hyperparameter $\sigma_n$ added to the diagonal.
The allowed operations are addition and multiplication.

The datasets used are generated by sampling ten datasets from four different \glspl{gp}.
These \glspl{gp} are standard initializations of GPyTorch \glspl{gp} with kernels Linear, SE, Matérn 3/2 and SE+SE.
For each of these $40$ datasets, a kernel search was performed, for each metric.
We generated datasets with 5, 10, 20, 30, 40, 50, 100 and 200 datapoints.

The approximated model evidence $\mathcal{Z}$ was calculated for the resulting kernel chosen by \gls{cks}.

For the calculation of the likelihood for test datasets, we sample another ten datasets from the same \glspl{gp} used to generate the training data.
This ensures, that they come from the same distribution.
For each such dataset we then calculate the \gls{mll} by replacing the training dataset with the test dataset, without retraining the model.

\subsubsection{Mauna Loa experiment}
In this setting we added an additional stopping criterion for the dynamical nested sampling, stopping after at most three million (3000000) samples to cap runtime.

\subsection{Choosing the hyperparameter prior}\label{app:prior_definition}
The normal distributed prior was selected by training \glspl{gp} with various kernels on randomly generated data from various \glspl{gp}.
Training was performed on data normalized to zero mean and one standard deviation, to make it consistent for the experiments.
All \glspl{gp} were trained on data from all other \glspl{gp}, e.g.\ a periodic kernel trained on data generated by a Matérn kernel, squared exponential kernel, etc. 
Hyperparameters are stored in their raw form, i.e.\ before applying the \verb|Positive| constraint from GPyTorch.
In particular, the hyperparameter priors are applied to the raw form.
The priors for the scaling hyperparameter $\sigma_f$ and the noise $\sigma_n$ are based on training results of \emph{all} kernels.
Show the hyperparameter priors for the raw values in Table~\ref{tab:param_priors}.
\begin{table}
    \centering
    \caption{The determined means and standard deviations for the hyperparameters of used kernels.}
    \label{tab:param_priors}
    \begin{tabular}{llll}
        \toprule
        Kernel & Hyperparameter & Mean & Standard deviation \\
        \midrule
        squared exponential & lengthscale & -0.212 & 1.89\\
        Matérn $3/2$ & lengthscale & 0.8 & 2.15 \\
        \multirow{2}{*}{periodic} & lengthscale & 0.78 & 2.29 \\
                                & period length & 0.65 & 1.0 \\
        \multirow{2}{*}{rational quadratic} & lengthscale & -0.05 & 1.94 \\
                                & alpha & 1.88 & 3.1 \\
        linear & variance & -0.8 & 1.0 \\
        scale kernel & $\sigma_f$ & -1.63 & 2.26 \\
        noise & $\sigma_n$ & -3.52 & 3.58 \\
    \end{tabular}
\end{table}
All our priors are normal distributions with diagonal covariances.
To construct the prior $p(\theta)\sim\mathcal{N}(\theta_\mu, \Sigma)$ for a specific \gls{gp} with kernel $k$ we concatenate the respective means and squared standard deviations from Table~\ref{tab:param_priors} to construct the vector $\theta_\mu$ and the diagonal matrix $\Sigma$.

The choice of these hyperparameter priors is \emph{not} the main reason for the superiority of our variants of the Laplace approximation over AIC and BIC, as otherwise MAP would benefit from these hyperparameter priors.
Clearly, the experiments show, that MAP is not a good model selection criteria.

\subsection{Laplace approximation derivations}
We derive the various Laplace approximations for the model evidence $p(y | X) = \int_\theta p(y | X, \theta)\cdot p(\theta) \operatorname{d}\!\theta$ with likelihood $p(y|X, \theta)$ and prior $p(\theta)$ for \gls{gp} hyperparameters $\theta$.

\subsubsection{Lap}\label{app:solution_model_evidence_integral_with_laplace_approximation}
We start with the application of the standard Laplace approximation.
\begin{align*}
    f(\theta) &= p(y | X, \theta)\cdot p(\theta) \\
    \log(f(\theta)) &= \log(p(y | X, \theta))\cdot p(\theta) \\
    \log(f(\theta)) &\approx \log(f(\hat{\theta})) - \frac{1}{2}(\theta-\hat{\theta})H(\theta-\hat{\theta})^T \\
    f(\theta) &\approx f(\hat{\theta}) \exp\left(-\frac{1}{2}(\theta-\hat{\theta})H(\theta-\hat{\theta})^T\right) \\
\end{align*}
where $H = -\nabla\nabla \log f(\theta)$.
We then insert the approximation into the model evidence integral.
\begin{align}
    p(y | X) &= \int_\theta f(\theta)\operatorname{d}\!\theta \\
     &\approx f(\hat{\theta})\cdot\int_\theta \exp\left(-\frac{1}{2}(\theta-\hat{\theta})H(\theta-\hat{\theta})^T\right)\operatorname{d}\!\theta \\
     &= f(\hat{\theta})\cdot\frac{(2\pi)^{\frac{-u}{2}}\cdot |H^{-1}|^{-\frac{1}{2}}}{(2\pi)^{\frac{-u}{2}}\cdot |H^{-1}|^{-\frac{1}{2}}}\int_\theta \exp\left(-\frac{1}{2}(\theta-\hat{\theta})H(\theta-\hat{\theta})^T\right)\operatorname{d}\!\theta \\
     &= f(\hat{\theta})\cdot\frac{1}{(2\pi)^{\frac{-u}{2}}\cdot |H^{-1}|^{-\frac{1}{2}}}\int_\theta (2\pi)^{\frac{-u}{2}}\cdot |H^{-1}|^{-\frac{1}{2}}\exp\left(-\frac{1}{2}(\theta-\hat{\theta})H(\theta-\hat{\theta})^T\right)\operatorname{d}\!\theta \\
     &= f(\hat{\theta})\cdot\frac{1}{(2\pi)^{\frac{-u}{2}}\cdot |H^{-1}|^{-\frac{1}{2}}} \cdot 1\\
     \log(p(y | X))&\approx \log(f(\hat{\theta})) + \frac{u}{2}\log(2\pi) + \frac{1}{2}\log(|H^{-1}|)\\
     \log(p(y | X))&\approx \widehat{\mathcal{L}}_{\text{MAP}} + \frac{u}{2}\log(2\pi) + \frac{1}{2}\log(|H^{-1}|)\\
     \log(p(y | X))&\approx \widehat{\mathcal{L}}_{\text{MAP}} + \frac{u}{2}\log(2\pi) - \frac{1}{2}\log(|H|) \label{eq:likelihood_approx_Lap}
\end{align}
where $u$ is the number of hyperparameters.
This leaves us with our approximation for the model evidence based on the $\log$ \gls{map} and the remaining correction term.

\subsubsection{Proof of Lemma~\ref{lem:laplace_approx_min_neg_contribution}}\label{app:laplace_minimal_negative_correction_derivation}
Due to the discussed pathologies of the Laplace approximation we replace some eigenvalues $\lambda$ of the Hessian $H$ with values based on minimal negative correction terms $r$.
We start with the approximation in formula \eqref{eq:likelihood_approx_Lap} and want to have a minimal negative correction value of $r$ for the term $+ \frac{u}{2}\log(2\pi) - \frac{1}{2}\log(|H|)$.
We motivated this approach during the discussion of Laplace approximations pathologies in section~\ref{subsection_pathologies}.
Again, in one dimension it holds that if a hyperparameter is superfluous, the likelihood surface will be mostly independent of it and, in the extreme, constant in its direction.
Thus, we correct the Hessian in those directions by replacing its eigenvalues.
\begin{proof}{\textit{Lemma~\ref{lem:laplace_approx_min_neg_contribution}}}
    Diagonalize the Hessian $H$ using the (orthogonal) matrix $T$ of eigenvectors to $\Lambda=THT^{-1}$, where $\Lambda$ is a diagonal matrix of eigenvalues.
    Then,  due to the determinant formula $|H| = |T||H||T^{-1}| = |\Lambda|$, we reduce to the case of a one-dimensional (i.e.\ $u=1$) inequality.
    Writing $\Lambda=\begin{bmatrix}\lambda\end{bmatrix}$ we get:
    \begin{align*}
    &&    r &\geq \frac{1}{2}\log(2\pi) - \frac{1}{2}\log(\lambda)\\
    &\iff&   2r &\geq \log(2\pi) - \log(\lambda)\\
    &\iff&   - 2r + \log(2\pi) &\leq \log(\lambda)\\
    &\iff&   \exp(-2r + \log(2\pi)) &\leq \lambda\\
    &\iff&   \exp(-2r)2\pi &\leq \lambda\\
    \end{align*}
	This is our intended result.
\end{proof}

\subsubsection{Proof of Lap collapsing into AIC/BIC}\label{app:Lap_collapse_proof}
Here we show that, in the case that all eigenvalues $\lambda$ of $H$ are small (i.e.\ they are all replaced with the lower bound), our approximation collapses into \gls{map} or \gls{aic} resp. \gls{bic} with \gls{map} as a surrogate for the likelihood function.
\begin{align*}
&&          \log(p(y|X)) &\approx \widehat{\mathcal{L}}_{MAP} + \frac{u}{2}\log(2\pi) - \frac{1}{2}\log(|H|)\\
&\iff&      \log(p(y|X)) &\approx \widehat{\mathcal{L}}_{MAP} + \frac{u}{2}\log(2\pi) - \frac{1}{2}\log(\prod_{i=1}^{u}2\pi\cdot\exp(-2r))\\
&\iff&      \log(p(y|X)) &\approx \widehat{\mathcal{L}}_{MAP} + \frac{u}{2}\log(2\pi) - \frac{1}{2}\sum_{i=1}^{u}\log(2\pi\cdot\exp(-2r))\\
&\iff&      \log(p(y|X)) &\approx \widehat{\mathcal{L}}_{MAP} + \frac{u}{2}\log(2\pi) - \frac{u}{2}\log(2\pi\cdot\exp(-2r))\\
&\iff&      \log(p(y|X)) &\approx \widehat{\mathcal{L}}_{MAP} + \frac{u}{2}\log(2\pi) - \frac{u}{2}\log(2\pi) + ur\\
&\iff&      \log(p(y|X)) &\approx \widehat{\mathcal{L}}_{MAP} + ur
\end{align*}

Now inserting our values for the minimal correction $r=0$, $r=-1$ and $r=-\log(n)$ we get back $\log(p(y|X)) \approx \widehat{\mathcal{L}}_{MAP}$, $\log(p(y|X)) \approx \widehat{\mathcal{L}}_{MAP} - u$ and $\log(p(y|X)) \approx \widehat{\mathcal{L}}_{MAP} - u\cdot\log(n)$, which are just $-\frac{1}{2}\text{AIC}$ resp. $-\frac{1}{2}\text{BIC}$ when choosing $\widehat{\mathcal{L}} = \widehat{\mathcal{L}}_{MAP}$ for \gls{aic} resp.\ \gls{bic}. 

\subsection{Licences}
GPyTorch \cite{gardner2018gpytorch} and Dynesty \cite{koposiv2023dynesty,speagle2020dynesty} are both licensed under the MIT license (\url{https://github.com/cornellius-gp/gpytorch/blob/master/LICENSE}, \url{https://github.com/joshspeagle/dynesty/blob/31b7e61031330ecb0b6df937b4013d8d592a6755/LICENSE}).

\subsection{Reproducibility}
For reproducibility we store the seeds used in the kernel search experiments.

\subsection{Compute}
\makeatletter
All training was performed on a server with an Intel(R) Core(TM) i9-10900K CPU @ 3.70GHz and 64GB of RAM.
\makeatother

\end{document}